\documentclass{article}

\usepackage{microtype}
\usepackage{graphicx}
\usepackage{subfigure}
\usepackage{booktabs} % for professional tables

\usepackage{hyperref}

\usepackage[arxiv]{icml2024c}

\usepackage{amsmath}
\usepackage{amssymb}
\usepackage{mathtools}
\usepackage{amsthm}

\usepackage[capitalize,noabbrev]{cleveref}

\theoremstyle{plain}

\theoremstyle{definition}

\theoremstyle{remark}

\usepackage{pifont}

\usepackage[textsize=tiny]{todonotes}

\icmltitlerunning{HEP-JEPA: JEPA-based foundation model for collider physics}

\begin{document}

\twocolumn[
\icmltitle{HEP-JEPA: A foundation model for collider physics using joint embedding predictive architecture}



\icmlsetsymbol{equal}{*}

\begin{icmlauthorlist}
\icmlauthor{Jai Bardhan}{yyy}
% {equal,yyy}
\icmlauthor{Radhikesh Agrawal}{equal,yyy}
% {equal,yyy,comp}
\icmlauthor{Abhiram Tilak}{equal,yyy}
% {comp}
\icmlauthor{Cyrin Neeraj}{yyy}
% {sch}
\icmlauthor{Subhadip Mitra}{yyy}
% {yyy}
\end{icmlauthorlist}

\icmlaffiliation{yyy}{International Institute of Information Technology, Hyderabad 500~032, India}

\icmlcorrespondingauthor{Jai Bardhan}{jai.bardhan90@gmail.com}

\icmlkeywords{Machine Learning, ICML}

\vskip 0.3in
]

\printAffiliationsAndNotice{\icmlEqualContribution} 

\begin{abstract}
We present a transformer architecture-based foundation model for tasks at high-energy particle colliders such as the Large Hadron Collider. We train the model to classify jets using a self-supervised strategy inspired by the Joint Embedding Predictive Architecture~\cite{assran2023self}. We use the JetClass dataset~\cite{https://doi.org/10.5281/zenodo.6619768} containing 100M jets of various known particles to pre-train the model with a data-centric approach --- the model uses a fraction of the jet constituents as the context to predict the embeddings of the unseen target constituents. Our pre-trained model fares well with other datasets for standard classification benchmark tasks. We test our model on two additional downstream tasks: top tagging and differentiating light-quark jets from gluon jets. We also evaluate our model with task-specific metrics and baselines and compare it with state-of-the-art models in high-energy physics. Project site: \href{https://hep-jepa.github.io/}{\url{https://hep-jepa.github.io/}}.
\end{abstract}

\section{Introduction}
Modern particle physics experiments at colliders, such as the Large Hadron Collider (LHC), rely heavily on deep learning models. Various crucial tasks such as tagging jets (collimated sprays of particles; roughly, tagging a jet means identifying the source elementary particle), reconstructing tracks of charged and neutral particles, matching detector measurements with simulations, etc., have seen many-fold improvements in performance with these sophisticated ML models. So far, the high-energy physics (HEP) community has primarily focused on building separate, dedicated deep-learning models for individual tasks using well-curated labelled simulated data. However, the current success of large world models~\citep{team2023gemini, reid2024gemini, garrido2024learningleveragingworldmodels} indicates that a `common knowledge' description could show better performance for individual tasks even in this case. 

Large Language Models (LLMs), such as the generative pre-trained transformer (GPT) models~\cite{brown2020languagemodelsfewshotlearners} and BERT~\cite{devlin2018pretraining}, have shown remarkable capabilities in learning generalised language representations by pre-training on vast amounts of texts. Similarly, a foundation model (FM) in HEP could learn to encode the `language' of particle interactions, detector responses, and physical laws, providing a versatile tool for analysing and interpreting experimental data. The success of LLMs indicates that FMs could revolutionise data-driven discovery in HEP, offering a scalable approach to uncovering new physics. The LHC has already started collecting massive amounts of data to probe new/rare processes. More data implies that the training time for models deployed at the collider will increase. Since FMs are pre-trained on huge amounts of data and perform well on downstream tasks with small fine-tuning, they can collectively save millions of compute hours for the HEP community. Researchers working with limited computational capabilities can also benefit from a FM for HEP.

Attempts to build FMs for HEP have started recently. With the available first-principle simulations in HEP, methods like OmniLearn~\cite{Mikuni:2024qsr} and Particle Transformer~\cite{Qu2022ParticleTF} rely on training transformer-based models in a supervised manner. Requiring labelled data, however, restricts their applicability to actual experimental data. Supervised training on simulated data can make models wrongly reliant on inherent limitations of the theoretical/numerical modellings (e.g., the absence of higher-order corrections or correlations). Hence, a self-supervised data-driven strategy is better suited to train a FM to learn from real experimental data. 

Masked Particle Modelling (MPM)~\cite{Golling:2024abg} and OmniJet-$\alpha$~\cite{Birk:2024knn}, which adapt Masked Language Modelling and Generative Pre-trained Transformers from the language domain, respectively, are examples where self-supervised learning (SSL) have been used to train FMs in HEP. We have also seen contrastive learning-based methods~\cite{Dillon:2021gag,Dillon:2022tmm,Dillon:2023zac} analogous to the \textsc{SimCLR} framework~\cite{simclr}. However, there are some limitations to the SSL approaches as well. For example, contrastive learning approaches require carefully chosen negative samples to avoid representational collapse~\cite{simclr}, and masked modelling methods often focus too much on local features, leading to less transferable representations. A major limitation of many SSL techniques comes from their reliance on decoders and reconstruction in the input space. The decoder often becomes the primary driver for learning, containing most of the model's inductive biases, resulting in weaker latent space representations, and an architecturally weak or constrained decoder can limit the quality of learned representations. Also, the computational overhead of maintaining and training a decoder can make these methods less efficient. Generative frameworks might be effective for their intended purpose, but they often learn details that are redundant for other tasks. For instance, precisely reconstructing every pixel or point might be superfluous for tasks like classification or object detection. These methods use computational resources to learn how to reproduce input-space details that might be irrelevant to semantic understanding.

In this paper, we train a FM for collider-related tasks within the Joint Embedding Predictive Architecture (JEPA) paradigm. Ref.~\citep{assran2023self} first proposed JEPA and demonstrated its effectiveness in image perception tasks. In contrast to other SSL strategies, JEPA learns predictive representations by modelling missing or unseen embeddings directly in the latent space without a decoder or full input reconstruction. By operating entirely in the representation space and eliminating the need for a decoder, JEPA becomes more computationally efficient and focuses on learning more abstract and transferable features. This approach allows the model to concentrate on capturing semantically meaningful patterns rather than being burdened by input-space reconstruction. The JEPA paradigm has shown significantly better performance than previous techniques for images~\citep{assran2023self}, videos~\citep{bardes2024revisiting}, and, recently, point clouds~\citep{saito2024point}.

Our contributions in this paper can be summarised as
\begin{enumerate}
    \item We adapt JEPA for HEP collider tasks and introduce a foundation model called HEP-JEPA. Most collider tasks require analysing scores of jets (highly challenging but necessary for physics experiments) produced in high-energy particle collisions. We use the JetClass dataset~\cite{https://doi.org/10.5281/zenodo.6619768} to pre-train our foundation model by providing parts of each jet sample as the context for the model to predict the rest of the jet correctly. 

    \item We present a framework to test model choices and HEP-JEPA performance with detailed ablations and comparisons with different training paradigms on few-shot learning tasks.
    
    \item We evaluate HEP-JEPA on two important downstream tasks: top tagging and differentiating light-quark jets from gluon jets. Top tagging is necessary for practically all new physics searches, and achieving a good discriminator of light jets is one of the significant challenges in the domain. We evaluate the model performance over reference datasets and task-specific metrics.   
\end{enumerate}

\section{JEPA pre-training paradigm}
JEPA learns to predict the embedding of a signal $y$ from a compatible signal $x$, using a predictor network conditioned on additional variables $z$ to facilitate prediction. Instead of predicting $y$ directly, JEPA predicts in representation space, which enables it to learn abstract meaningful representations of inputs, making it ideal for some downstream tasks. Our HEP-JEPA is one instantiation of the paradigm to work with particle jets by masking a fraction of jet constituents. Similar to the joint embedding architecture, JEPA is susceptible to representation collapse, which we bypass using an exponential-moving-average teacher design for the $y$ encoder.

\section{HEP-JEPA model}\label{sec:HEPJEPA_model}

\begin{figure*}
    \centering
    \includegraphics[width=\linewidth]{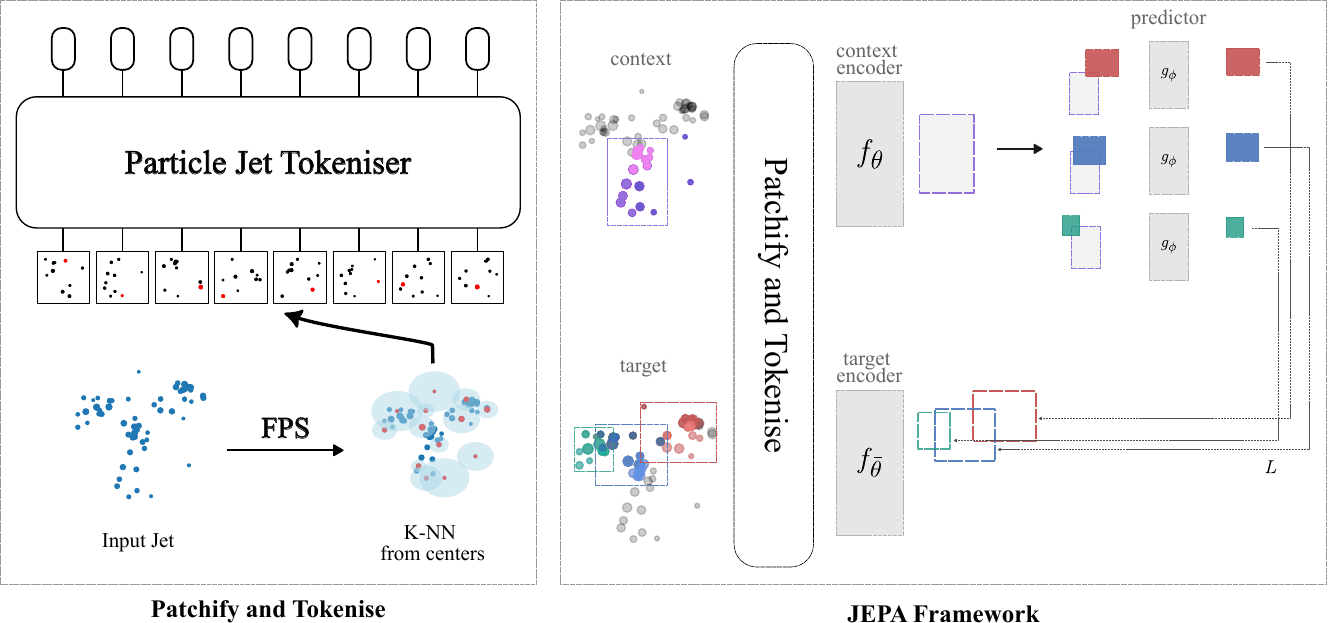}
    \caption{Schematic diagram illustrating the working of the HEP-JEPA model. The model has a structure similar to vision transformers. In the first step, the entire jet is divided into patches using a particle jet tokeniser. These tokens are then masked to form the context and target blocks. Each block is fed into the respective encoder to generate the embeddings. The context embedding, along with the special mask tokens, is used by the predictor to predict the embedding of the masked target blocks.}
    \label{fig:HEP-JEPA-schematic}
\end{figure*}

Figure~\ref{fig:HEP-JEPA-schematic} shows the complete HEP-JEPA framework, which works as follows. For every input set (jet) of particles (each with $2$ coordinates $\eta, \phi$ --- pseudorapidity and azimuthal angle --- and a set of kinematic variables), we first divide it into geometrical patches based on the coordinates of each particle. Each patch of the jet is encoded using a small permutation invariant model to form patch tokens. These tokens are then divided into non-overlapping context ($x$) and target ($y$) sets and encoded by their respective encoders. A predictor is used to predict the embedding of the target tokens. The encoders are implemented using a transformer encoder architecture.   

\subsection{Particle group tokeniser}
Let us consider a jet $J$ consisting of $n$ particles where each particle is represented by a vector $p_i \in \mathbb{R}^7$:
\begin{align*}
 p_i =(&\eta_i, \phi_i, m_i, \ln p_{T_i}, \nonumber \ln E_i, \ln \frac{p_{T_i}}{p_{T_J}}, \ln \frac{E_i}{E_J}, \Delta_{ R_{iJ}}),
\end{align*}
where $p_T$ denotes the momentum component transverse to the beam direction, $m$ denotes the mass, $E$ is the energy, and $\Delta_{ R_{iJ}}$ is the distance between the particle and the jet axis in the $(\eta,\phi)$ plane. We introduce a patchification and tokenisation strategy to capture meaningful \textit{patch-level} particle interactions. The tokeniser $T: \mathbb{R}^{n \times 7} \to \mathbb{R}^{c \times d}$ maps the raw particle features to $c$ token embeddings of dimension $d$.

First, we employ Farthest Point Sampling to select $c$ centre particles that maximise the jet's phase space coverage. We construct a local group of $k$ particles for each centre using the $k$-nearest neighbours in the $(\eta, \phi)$ plane. These groups are then normalised by subtracting their respective centre coordinates:
\[
p'_i = p_i - p_k, \quad \forall i \in G_k
\]
where $G_k$ denotes the group of particles associated with centre $k$. To obtain permutation-invariant token embeddings, we process each normalised group through a small PointNet encoder, $E$, consisting of shared multilayer perceptions (MLPs) followed by max-pooling operations:
\[
t_k = \max\left( \{ \text{MLP}(p'_i) \mid p'_i \in G_k \} \right).
\]
In our case, $c$ is not a fixed number as it varies with the size of the particle jet. We sample only a small fraction of the points as the possible centres for the jet. 

\subsection{JEPA framework}

\subsubsection{Token construction}

\paragraph{Target construction:}
Our masking strategy operates on the token sequence $T = \{t_1, \dots, t_k\}$ to create complementary context and target regions. The $c$ groups of the jet $J$ is fed through the target encoder $f_{\bar{\theta}}$ to obtain a corresponding patch-level embedding $\boldsymbol{s}_{y} = \{\boldsymbol{s}_{y_1}, \dots, \boldsymbol{s}_{y_c}\}$, where $\boldsymbol{s}_{y_k}$ is the representation associated with the $c^{\text{th}}$ group. For each jet, we sample $M$ (possibly overlapping) blocks from the target representations $\boldsymbol{s}_y$ with random scale $s \in [0.15, 0.2]$ and random aspect ratio $r \in [0.75, 1.5]$. As has been shown previously, it is essential to mask the output of the target encoder and not the input.

\paragraph{Context construction:}

The context block is sampled with a random scale $s \in [0.4, 0.75]$ from the set of all tokens. Since the target and the context are selected independently, they may overlap significantly. We remove overlapping regions between the source and target tokens to prevent trivial learning. These tokens are fed to the context encoder $f_{\theta}$ to obtain the corresponding representations $\boldsymbol{s}_x = \{s_{x_j} \}_{j \in B_x}$, where $B_x$ is the mask associated with block $x$. 

Following Point-JEPA~\citep{saito2024point}, we also utilise a greedy sequencing algorithm to ensure spatial coherence when performing the contiguous masking of source and target tokens. Below we summarise the main steps of the sequencer.
\begin{enumerate}
    \item Initialise with token $t_i$ minimizing $\sum_i \text{coord}(t_i)$.
    \item Iteratively select the next token based on the minimum distance on the $(\eta, \phi)$ plane.
    \item Maintain disjoint sets between context and target tokens.
\end{enumerate}

\subsubsection{Encoder architectures}
Our framework consists of three primary components operating in representation space:

\paragraph{Context and target encoders:} 
Our context and target encoders are transformer models for encoding the tokens in a jet sample. We use eight registers with each transformer encoder, following Ref.~\cite{darcet2024visiontransformersneedregisters}. We also introduce a physics bias matrix from Ref.~\cite{Qu2022ParticleTF}; however, our implementation differs since we calculate pairwise interactions between tokens (groups) instead of particles. This requires calculating the four-momentum vectors of the entire group. The bias terms are:
\begin{gather*}
    \Delta_{R_{ij}} = \sqrt{(\eta_i - \eta_j)^2 + (\phi_i - \phi_j)^2}, \\ 
    k_{T} = \min(p_{T_i}, p_{T_j})\Delta_{R_{ij}}, \\
    z = \min(p_{T_i}, p_{T_j}) / (p_{T_i} + p_{T_j}), \\
    m^2 = (E_i + E_j)^2 - \|\mathbf{p}_i + \mathbf{p}_j\|^2,
\end{gather*}
where $\mathbf{p}_i$ denotes the momentum of the $i^{\text{th}}$ particle. The bias for the added registers is set to $0$.

\paragraph{Predictor:}
The predictor predicts the representations of targets with the help of context blocks. For a given target block $\boldsymbol{s}_y(i)$ corresponding to a target mask $B_i$, the predictor $g_\phi(\cdot, \cdot)$ takes the context encoder $\boldsymbol{s}_x$ and a mask token for each patch we wish to predict $\{\boldsymbol{m}_j\}_{j \in B_i}$ as inputs, and outputs a patch-level prediction $\hat{\boldsymbol{s}}_y(i) = \{\hat{\boldsymbol{s}}_{y_j}\}_{j \in B_i} = g(\boldsymbol{s}_x, \{\boldsymbol{m}_j\}_{j \in B_i})$. The mask tokens are parametrised by a shared learnable vector with an added positional embedding. Since we want to predict for each of the $M$ blocks, we apply the predictor $M$ times.

\paragraph{Loss function:}
The learning objective is formulated entirely in the embedding space as
\[
L = \frac{1}{M} \sum_{i=1}^{M} D(\hat{s}_y^{(i)}, s_y^{(i)}),
\]
where $D$ is the smooth L1 loss between predicted embeddings $\hat{\boldsymbol{s}}_y^{(i)} = g(\boldsymbol{s}_x, \{\boldsymbol{m}_j\}_{j \in B_i})$ and target embeddings $\boldsymbol{s}_y^{(i)} = \{\boldsymbol{s}_{y_j}\}_{j \in B_i}$. By operating in the representation space rather than the particle space, we focus the learning on physically relevant features while avoiding the computational overhead of full reconstruction.

\section{Pre-training HEP-JEPA}
We pre-train our models on the JetClass dataset~\cite{https://doi.org/10.5281/zenodo.6619768}. JetClass is an extensive collection of jets clustered from simulated proton-proton collisions at the LHC. It contains $100$M training samples divided into $10$ jet classes, covering light jets (from gluons and light quarks) and heavy-particle jets from the top quark or the Higgs, $W$, and $Z$ bosons. Each jet is reconstructed using the anti-$k_T$ algorithm~\cite{Cacciari:2008gp} (with jet radius $R=0.8$) after incorporating detector effects from \textsc{Delphes}~\cite{deFavereau:2013fsa}, the detector simulator. The dataset provides detailed per-particle features, including kinematics (energy-momentum four-vectors), particle identification ($5$-class encoding), and trajectory displacement parameters for tagging heavy-flavor jets. It is split into training ($100$M jets), validation ($5$M), and test ($20$M) sets.

\paragraph{The model and its training:}
For the model, we take a standard transformer architecture of $12$ transformer blocks with $2.5$M parameters in total. We train the model for $4$ epochs with an effective batch size of $2048$, corresponding to roughly $200$k training steps. We take a cosine decay scheduler with a linear warm-up for $15$k steps. Our training was complete in $320$ GPU hours on RTX 2080Ti.

\section{Experiments}\label{sec:experiments}

\subsection{Evaluations on JetClass dataset}
\subsubsection{Few-shot learning}
We perform a few-shot evaluations on the JetClass dataset. As the baseline model, we take our architecture trained from scratch in a supervised fashion to eliminate the effects of other external factors. In both these models (ours and the baseline), we attach a classification head to the student backbone constructed as two class attention blocks followed by a MLP layer. We consider two regimes to perform the few shot evaluation: 
\begin{enumerate}
    \item[(1)] \textit{frozen:} where the pre-trained backbone is not updated -- only the classification head is trained, and 
    \item[(2)] \textit{fine-tuned:} where the pre-trained backbone is simultaneously updated with the classification head.
\end{enumerate}
We conduct the experiment at different label fractions of the JetClass dataset -- $0.05\%$ $0.5\%$, $2\%, 10\%$, and $100\%$.
%%%%%%%%%%%%%%%%%%%%%%%%%%%%%%%%%%%%%%%%%%%%%%%%%%%%%%%%%%%%%%%
\begin{table}[h!]
\caption{JetClass Metrics: Peak validation accuracies attained by the benchmark models on the JetClass dataset. We run experiments for different levels of few-shot learning. We provide different fractions of labels from the JetClass dataset to train the model from scratch for the classification task. In each experiment, a pre-trained HEP-JEPA model is fine-tuned on the same fraction of labels.}
\label{tab:res_jetclass}
\vskip 0.15in
\begin{center}
\begin{small}
\begin{sc}
\begin{tabular}{clr}
\toprule
\% of labels &Model & Accuracy\\
\midrule
0.05\% & From scratch & $0.505$ \\
(5k) & HEP-JEPA, fine-tuning & $0.564$ \\
\midrule
0.5\% & From scratch & $0.586$ \\
(50k) & HEP-JEPA, fine-tuning & $0.624$ \\
\midrule
2\% & From scratch & $0.668$\\
(2M)  & HEP-JEPA, fine-tuning & $0.669$ \\
\midrule
10\% & From scratch & $0.683$ \\
(10M) & HEP-JEPA, fine-tuning & $0.685$ \\
\midrule
100\% & From scratch & $0.698$ \\
(100M) & HEP-JEPA, fine-tuning & $0.698$ \\
\bottomrule
\end{tabular}
\end{sc}
\end{small}
\end{center}
\vskip -0.1in
\end{table}
%%%%%%%%%%%%%%%%%%%%%%%%%%%%%%%%%%%%%%%%%%%%%%%%%%%%%%%%%%%%%%%%%%
\begin{table}[h!]
\caption{\emph{Downstream task I - top tagging}: Peak validation accuracies for the benchmark models on the TQTR dataset~\cite{kasieczka_2019_2603256} using $100\%$ of the data samples. Accuracies of two benchmark models, one trained from scratch on the TQTR dataset and the pre-trained HEP-JEPA model, are shown for different fine-tuning levels on the TQTR dataset. ``\textsc{Frozen}'' after a model name indicates that the pre-trained was not updated while training, whereas the tag ``\textsc{Fine-tuned}'' indicates that both the classifier head and the pre-trained backbone were updated during the training. We also report the performances of two state-of-the-art models on top tagging from Refs.~\cite{Qu:2019gqs,Qu2022ParticleTF}, indicated with (*), for a comparison to the HEP-JEPA performance.}
\label{tab:res_toptag}
\vskip 0.15in
\begin{center}
\begin{small}
\begin{sc}
\begin{tabular}{lr}
\toprule
Model & Accuracy\\
\midrule
From scratch & $0.927$\\
Supervised, Frozen & $0.928$\\
Supervised, Fine-tuned & $0.938$\\
\midrule
HEP-JEPA, Frozen & $0.928$ \\
HEP-JEPA, Fine-tuned & $0.929$ \\
\midrule
ParticleNet (*) & $0.940$ \\
ParT (*) & $0.944$ \\
\bottomrule
\end{tabular}
\end{sc}
\end{small}
\end{center}
\vskip -0.1in
\end{table}

\begin{figure}
    \centering
    \includegraphics[width=0.9\columnwidth]{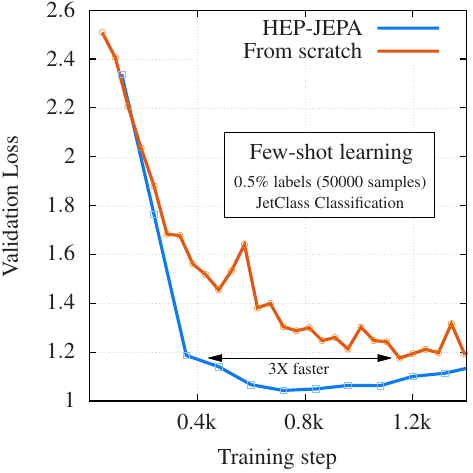}
    \caption{Validation loss vs. training step for the two benchmark models training in a few-shot learning setting for jet classification on the JetClass dataset with $0.5\%$ labels (i.e., $50000$ training samples). One model is trained from scratch, whereas the pre-trained HEP-JEPA model is fine-tuned. The validation loss falls quickly for the HEP-JEPA model --- it achieves the same minimum validation loss as the model trained from scratch three times faster.}
    \label{fig:valLoss-50kLabels}
\end{figure}

Table~\ref{tab:res_jetclass} shows the macro accuracy obtained by the methods. The fine-tuned HEP-JEPA model consistently outperforms the model trained from scratch. The difference is significant in few-shot learning tasks (i.e., $0.05\%$- and $0.5\%$-label cases), where HEP-JEPA shows $4$ $-$ $6$\% better accuracies than the model trained from scratch. To illustrate this, we show Figure~\ref{fig:valLoss-50kLabels}, where we plot the validation loss against the training step for the two benchmark models training in a few-shot learning setting for jet classification on the JetClass dataset with $0.5\%$ labels. However, as the fraction of labels for training increases, the performance difference between the benchmarks reduces. HEP-JEPA performs almost identically to the model trained from scratch when the complete set of labels is available for training.

\subsection{Transfer learning evaluations on downstream tasks}
We also test the model's capabilities for generalising and its performance on different datasets.

\subsubsection{Top tagging}
The top quark is the heaviest known particle of the Standard Model. Once produced, it quickly decays inside the collider and predominantly produces a three-pronged jet (from its hadronic, i.e.,  $t\to W b \to q q^\prime b$ (three-body) decays). The dominant background processes %(e.g., light jets) 
also often produce similar jet signatures, making it challenging to achieve high signal-to-background ratios. To evaluate how HEP-JEPA performs in tagging top jets, we use the Top Quark Tagging Reference (TQTR) dataset~\cite{kasieczka_2019_2603256}, which consists of 2M samples of jets originating from hadronic decays of the top quark as well as ones from lighter quarks and gluons, with the recommended split of $1.2$M samples for training, $400$k each for validation and testing.

We evaluate the performance of our model architecture with two benchmark models: one trained on the JetClass dataset in a supervised fashion and the other trained on the TQTR dataset from scratch. We also compare how fine-tuning affects the performance of these benchmark models.

Table~\ref{tab:res_toptag} shows the top-tagging metrics %for the classification tasks 
obtained using $100\%$ of the dataset. The fine-tuned versions of the benchmark models perform better than their frozen versions. The fine-tuned HEP-JEPA attains a better accuracy score than both the supervised model trained from scratch and the frozen HEP-JEPA. However, we see that the fine-tuned supervised model shows better performance than the other benchmarks. To understand this, we should keep in mind that 1) the kinematic range of jet samples from the TQTR dataset (transverse momenta $p_T \in \left[550, 650\right]$) is smaller than that of the JetClass dataset ($p_T \in \left[500, 1000\right]$) and 2) HEP-JEPA is trained on the auxiliary task predicting missing parts of the jet samples, whereas the supervised model is trained for the particular classification task. Both factors could give a slight edge to the fine-tuned supervised benchmark model. We also show the accuracy scores from the domain state-of-the-art models~\cite{Qu:2019gqs,Qu2022ParticleTF} for reference.

\begin{table}[t]
\caption{\emph{Downstream task II - quark-gluon tagging}: Peak validation accuracies for the benchmark models on a reference quark-gluon tagging dataset~\cite{komiske_2019_3164691} using $100\%$ of the data samples. Accuracies of the two benchmark models, one trained from scratch on the dataset and the pre-trained HEP-JEPA model, are shown. The tag ``\textsc{Frozen}'' after a model name indicates that the pre-trained model backbone was not updated during the finetuning. We also report the performances of two state-of-the-art models on top tagging from Refs.~\cite{Qu:2019gqs,Qu2022ParticleTF}, indicated with (*), for a comparison to the HEP-JEPA performance.}
\label{tab:res_qgtag}
\vskip 0.15in
\begin{center}
\begin{small}
\begin{sc}
\begin{tabular}{lr}
\toprule
Model & Accuracy\\
\midrule
From scratch & $0.819$ \\
Supervised, Frozen & $0.823$\\
\midrule
HEP-JEPA, frozen & $0.821$\\
\midrule
ParticleNet (*) & $0.840$ \\
ParT (*) & $0.843$ \\
\bottomrule
\end{tabular}
\end{sc}
\end{small}
\end{center}
\vskip -0.1in
\end{table}

\subsubsection{Quark vs. gluon jet tagging}
Accurately tagging light jets is one of the most important open problems in collider physics --- it is not yet possible to make out (at least reliably) whether a light jet originated in a light quark ($u$, $d$, or $s$) or a gluon at the LHC. The ability to do so would open a new window to new physics searches and significantly enhance the sensitivity of rare process searches. To evaluate how HEP-JEPA perform on this issue, we use the quark-gluon tagging dataset~\cite{komiske_2019_3164691} containing $2$M samples of quark and gluon jets modelled using \textsc{Pythia}~\cite{Sjostrand:2014zea} without any detector effects.

As in the case of top tagging, we evaluate the performance of two benchmark models, i.e., a fully supervised model and a model trained using JEPA on the JetClass dataset for this task.

Table~\ref{tab:res_qgtag} shows the accuracies of quark-gluon tagging when the models are trained with $100\%$ of the samples. The numbers show a similar trend to the top tagging case. The fine-tuned HEP-JEPA outperforms the model trained from scratch but slightly falls short of when the latter is further fine-tuned on the dataset. We also show the accuracy scores from the domain state-of-the-art models~\cite{Qu:2019gqs,Qu2022ParticleTF} for reference.

\subsection{Visualisation}
\begin{figure}
    \centering
    \includegraphics[width=\linewidth]{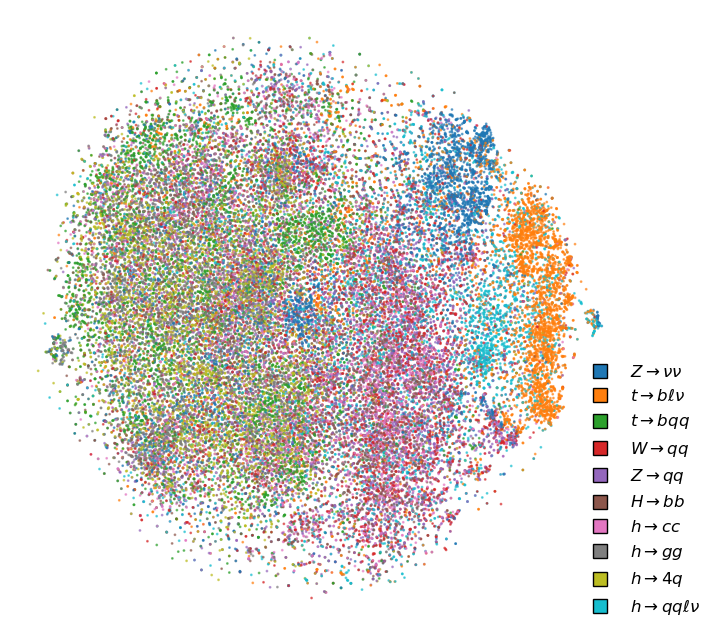}
    \caption{The t-SNE plot of the pooled embedding obtained for samples within the JetClass dataset.}
    \label{fig:jetclass_tsne}
\end{figure}

We visualise the representation learned by HEP-JEPA on $50$k samples of JetClass sampled uniformly from each class. We construct the embedding for a sample by concatenating the max and mean pooling of the outputs of the context encoder and apply t-SNE on the pooled embedding. We visualise the results in Figure~\ref{fig:jetclass_tsne}. We observe that events that contain lepton(s) are pushed to the right, while hadronic events are more towards the left.

\section{Ablation tests}
We perform ablations on our model choices to understand how they affect the model performance. Our ablation test setup is as follows. For all tests, we pre-train our model for an epoch (roughly $48000$ training steps) on the $100$M training samples of the JetClass dataset. We estimate the final performance using a SVM linear classifier trained on $50$k training samples and evaluate on $50$k validation samples obtained by stratified random sampling. Given our resource constraints, we make these reductions to allow us to prototype ablation decisions quickly. We categorise and describe the ablation studies in detail below.

\subsection{Ablation tests on JEPA framework parameters}

\paragraph{Context and Target selection (masking) strategy:}
We test two strategies for target masking --- \textit{random} and \textit{contiguous} selection. In \textit{random} selection, we randomly select a fraction of target tokens from all the tokens to mask. In \textit{contiguous} selection, we select a fraction of tokens corresponding to neighbouring regions in the $(\eta,\phi)$ plane. The latter is implemented using the Point Sequencer from Ref.~\cite{saito2024point}. We also run a preliminary investigation on the effect of different context sample ratios.

\paragraph{Number of targets:}
We also test the number of target tokens that the model needs to predict for each context token; specifically, we check with $1$, $4$, and $8$ target tokens. The results indicate that the number of target tokens does not significantly influence performance. 
\begin{table}[h]
\caption{Ablation I: Peak validation accuracies for the number of targets and masking strategies. The context sample ratio fixed to the range $\left[0.15, 0.2\right]$ and the target sample ratio to $\left[(0.4, 0.75\right]$.} 
\label{tab:abls_target}
\vskip 0.15in
\begin{center}
\begin{small}
\begin{sc}
\begin{tabular}{lcr}
\toprule
Strategy & Frequency & Accuracy \\
\midrule
Random & 4 & $0.579$\\
Contiguous & 1 & $\mathbf{0.581}$\\
Contiguous & 4 & $0.557$\\ % VANILLA CASE
Contiguous & 8 & $0.562$\\
\bottomrule
\end{tabular}
\end{sc}
\end{small}
\end{center}
\vskip -0.1in
\end{table}

\begin{table}[h]
\caption{Ablation test II: Peak validation accuracies for different context sample ratio. The target sample ratio is in the range (0.15, 0.2) with the number of targets set to 1.} 
\label{tab:abls_context}
\vskip 0.15in
\begin{center}
\begin{small}
\begin{sc}
\begin{tabular}{lcr}
\toprule
Strategy & Context Sample Ratio & Accuracy \\
\midrule
Contiguous & $(0.4, 0.75)$ & $0.557$\\ % VANILLA CASE
Contiguous & $(0.85, 1.0)$ & $\mathbf{0.563}$\\
\bottomrule
\end{tabular}
\end{sc}
\end{small}
\end{center}
\vskip -0.1in
\end{table}

From these tests, we find that a contiguous target masking strategy with one target is a better model choice. After the target is selected, a sample ratio in the range $\left[ 0.85, 1.0\right]$ performs better.

\subsection{Ablation tests for model architecture choices}

\paragraph{Physics bias for the attention mechanism:}
We also analyse the impact of physics bias on the model's performance. Since each token corresponds to a group of particles, we calculate the pairwise bias terms for groups of particles by first summing the four-vectors of the particles within a group to get the group-level four-vector. Table~\ref{tab:abls_phybias} show that the model performs $\approx2\%$ better after including the physics bias. 
% \begin{gather*}
% \end{gather*}
\begin{table}[h!]
\caption{Ablation III: Peak validation accuracies with/without physics bias for the attention mechanism. } 
\label{tab:abls_phybias}
\vskip 0.15in
\begin{center}
\begin{small}
\begin{sc}
\begin{tabular}{cc}
\toprule
Physics Bias & Accuracy \\
\midrule
\ding{52} & $\mathbf{0.570}$ \\ % CHECK FONT HERE!!!
\ding{53} & $0.557$ \\ % CHECK FONT HERE!!!
\bottomrule
\end{tabular}
\end{sc}
\end{small}
\end{center}
\vskip -0.1in
\end{table}

\paragraph{Integrating registers with our transformer model:}
Adding additional tokens to the input sequence of the Vision Transformers has been shown to yield smoother feature maps and attention maps for downstream visual processing~\cite{darcet2024visiontransformersneedregisters}. Given the similarity of our transformers with vision transformers, we investigate the impact of integrating register tokens with our transformer blocks.  Table~\ref{tab:abls_reg} shows that the model performance increases by $\approx2\%$ with registers.
\begin{table}[h]
\caption{Ablation IV: Peak validation accuracies with/without integrating registers with our transformer model.} 
\label{tab:abls_reg}
\vskip 0.15in
\begin{center}
\begin{small}
\begin{sc}
\begin{tabular}{cc}
\toprule
Registers & Accuracy \\
\midrule
\ding{52} & $\mathbf{0.576}$\\
\ding{53} & $0.557$\\
\bottomrule
\end{tabular}
\end{sc}
\end{small}
\end{center}
\vskip -0.1in
\end{table}

\subsection{Data preprocessing ablation tests}

\paragraph{Physics-inspired augmentations:}
We also perform a preliminary study of the impact of adding physics-based data augmentations to the jets. Particularly, we test a combination of rotation, smearing, and boosting. However, we do not find any significant improvement in the model's performance by training it on the augmented data. 
\begin{table}[h]
\caption{Ablation V: Peak validation accuracies with and without physics augmentations.}
\label{tab:abls_phyaugs}
\vskip 0.15in
\begin{center}
\begin{small}
\begin{sc}
\begin{tabular}{cc}
\toprule
Augmentations & Accuracy \\
\midrule
\ding{52} & $0.553$\\
\ding{53} & $0.557$\\
\bottomrule
\end{tabular}
\end{sc}
\end{small}
\end{center}
\vskip -0.1in
\end{table}

Table~\ref{tab:abls_phyaugs} shows that this test is inconclusive --- the model performance is similar in both cases.

\section{Related works}\label{sec:relatedwork}
The development of foundation models has revolutionised the field of artificial intelligence, enabling generalised learning across diverse tasks and domains~\cite{bao2022beitbertpretrainingimage,DBLP:journals/corr/abs-2104-14294,kim2024openvlaopensourcevisionlanguageactionmodel,touvron2023llamaopenefficientfoundation}. Various paradigms for developing foundational models have been introduced, including generative approaches like Masked Autoencoders (MAE)~\cite{he2021maskedautoencodersscalablevision} and its extension to 3D point clouds in Point-MAE~\cite{pang2022maskedautoencoderspointcloud}, multi-scale masked autoencoders such as Point-M2AE~\cite{zhang2022pointm2aemultiscalemaskedautoencoders}, and contrastive learning methods exemplified by SimCLR~\cite{chen2020simpleframeworkcontrastivelearning}.  JEPA~\cite{assran2023self} is a novel training paradigm for learning effective abstract representations while addressing the shortcomings of other self-supervised models. The framework was initially proposed for images and has been adapted to the domains of videos~\citep{bardes2024revisiting} and, recently, to point clouds~\citep{saito2024point}. 

The success of these large models in computer science has led to the development of similar models in fundamental sciences like biology~\cite{doi:10.1073/pnas.2016239118, ross2022largescalechemicallanguagerepresentations,doi:10.1021/acs.jcim.4c01396}, chemistry~\cite{liao2024wordsmoleculessurveylarge,Irwin_2022}, astronomy~\cite{Parker_2024} and the modelling of dynamical systems in general~\cite{subramanian2023foundationmodelsscientificmachine, mccabe2024multiplephysicspretrainingphysical}. In the domain of HEP, Masked Particle Modelling (MPM)~\cite{Golling:2024abg, Leigh:2024ked} was one of the first attempts to build such a model. It uses a self-supervised strategy to train on the JetClass dataset, drawing inspiration from language models. The MPM method models the jets as unordered sets of particles, each attributed with continuous features such as momenta, point of detection within the detector, etc. For a given jet, a subset of the particles are ``masked'' (i.e., their information is removed and replaced with a learnable vector or ``mask'') and fed to a transformer model. The model has to predict the masked part correctly using the information of the unmasked part to recreate the representation of the jet. Another language models-inspired method, OmniJet-$\alpha$~\cite{Birk:2024knn}, uses a transformer-based model to generate tokenised jets from the JetClass dataset in an autoregressive manner similar to the GPT models~\cite{brown2020languagemodelsfewshotlearners}. The authors show that the model generalises well to a supervised classification task on the same set of jets, especially in few-shot learning, where the fine-tuned model outperforms the model architecture trained for the same task from scratch. Other people have also used contrastive learning techniques ~\cite{Dillon:2021gag,Dillon:2022tmm,Dillon:2023zac,Harris:2024sra}. The OmniLearn method~\cite{Mikuni:2024qsr} relies on robust first-principle HEP simulations to train a FM in a supervised manner. The model has a classifier head and a generator head to cover most learning tasks, forming a task-specific block. This task-specific block uses a shared representation learned by the point-edge transformer model while training on a jet classification task on the \textsc{JetClass} dataset. The model matches the benchmark supervised jet classification model, Particle Transformer (ParT)~\cite{Qu2022ParticleTF}. The pre-trained model is evaluated across multiple learning tasks -- it trains quicker while performing comparably to the state-of-the-art on these tasks. 

Concurrent to our work, Ref.~\cite{katel2024learningsymmetryindependentjetrepresentations} adapts the JEPA paradigm for the task of top tagging --- the authors pre-train the model on $1\%$ of the top jet and light jet samples from JetClass and evaluate downstream performance on the TQTR dataset. However, unlike our data-centric approach, the authors provide a physics-motivated context for the predictor -- a subset of the \emph{subjets} created from a jet sample from which it has to learn the target \emph{subjet} embeddings. Our model attains better accuracies for top tagging. Moreover, we also perform model evaluations for an additional downstream task of quark-gluon tagging, along with detailed ablations.

Building HEP-JEPA is the first attempt to thoroughly evaluate the adoption of the JEPA paradigm to the high-energy physics domain with different model-building choices.

\section{Conclusions}
High-energy physics is a data-intensive and experimentally challenging domain, where a successful foundation model can accelerate the search for new physics and lead to fundamental insights in physics. Particularly in jet physics, foundation models can effectively capture inherent complex patterns and reshape how we approach challenging tasks in collider physics, thereby reducing the use of computational resources significantly. In this paper, we introduced the JEPA framework to the domain of HEP and showed its capabilities with two crucial downstream tasks: top tagging and quark-gluon tagging. We also showed its effectiveness in few-shot learning settings for jet classification tasks. 

The JEPA paradigm was tested and thoroughly evaluated on an extensive HEP dataset for the first time. Though its current performance indicates scopes for improvement compared to other state-of-the-art task-specific methods like ParT~\cite{Qu2022ParticleTF} (in the absolute sense), JEPA-based training has the ability to learn better abstract representations and shows superior scalability to real experimental data. These reasons impel us to improve HEP-JEPA further.

JEPA as a paradigm is also largely independent of the underlying model backbone and, therefore, can benefit from improvements to the underlying model architecture. Evaluating HEP-JEPA on other tasks, such as unfolding detector measurements to improve first-principle simulations and anomaly detection using weakly supervised methods, could further validate its generalisability [this can be easily done with publically available datasets~\cite{andreassen_2019_3548091} and~\cite{kasieczka_2019_4536624} respectively]. Additionally, extending HEP-JEPA to generative tasks and event classification is the next step to make it a truly cross-task FM for collider physics. 

\section*{Impact statement}

Our goal is to accelerate particle physics research through Machine Learning. We do not foresee our work directly introducing any societal risks; on the contrary, we believe this could benefit society in terms of energy savings, democratisation of research, etc. For example, much time and computational resources go into simulating and analysing collision events. A pre-trained foundational model would enable the rapid development of task-specific models through fine-tuning on smaller datasets, thereby reducing the computational overhead of training models and generating data.

\bibliography{FMJEPA}
\bibliographystyle{icml2024}

\end{document}